\documentclass{article}

\usepackage[final]{corl_2018} 

\usepackage{times}
\usepackage{epsfig}
\usepackage{graphicx}
\usepackage{float}
\usepackage{wrapfig}

\usepackage{amsmath,amssymb}

\usepackage{bm,xspace}
\usepackage{comment}
\usepackage{verbatim}
\usepackage{multirow}
\usepackage{balance}
\usepackage{url}
\usepackage{booktabs}
\usepackage{etoolbox,siunitx}
\usepackage{calc}
\usepackage{pifont,hologo}
\usepackage{color}

\usepackage[super]{nth}
\usepackage{nicefrac}
\def\aa{\mathbf{a}}

\def\cc{\mathbf{c}}

\def\oo{\mathbf{o}}

\def\tt{\mathbf{t}}

\DeclareMathOperator*{\argmin}{arg\,min}

\def\deg{^{\circ}}

\DeclareMathSymbol{@}{\mathord}{letters}{"3B}

\newcommand\norm[1]{\left\lVert#1\right\rVert}
\newcommand\tuple[1]{\left\langle#1\right\rangle}

\newcommand\timess{\mathbin{\!\times\!}}

\definecolor{alexey}{rgb}{0.8,0.,0.8}

\newcommand\mypara[1]{\noindent\textbf{#1}}

\def\latex/{\LaTeX}
\def\bibtex/{\hologo{BibTeX}}

\newcommand{\figLabel}{Figure~}

\usepackage{capt-of}
\usepackage{rotating}
\usepackage{multirow}
\usepackage{pgfplots}

\newcommand{\act}{\aa}

\newcommand{\obs}{\oo}

\newcommand{\cmd}{\cc}

\newcommand{\params}{\boldsymbol{\theta}}

\newcommand{\cmdleft}{{\texttt{left}}}
\newcommand{\cmdstraight}{{\texttt{straight}}}
\newcommand{\cmdright}{{\texttt{right}}}

\graphicspath{{./images/}}

\usepackage{times}

\usepackage[numbers]{natbib}
\usepackage{multicol}

\title{Driving Policy Transfer via Modularity and Abstraction}

\author{
  Matthias M\"uller\\
  Visual Computing Center\\
  KAUST, Saudi Arabia\\
  \And
  Alexey Dosovitskiy \\
  Intelligent Systems Lab \\
  Intel Labs, Germany \\
  \AND
  Bernard Ghanem \\
  Visual Computing Center\\
  KAUST, Saudi Arabia\\
  \And
  Vladlen Koltun \\
  Intelligent Systems Lab \\
  Intel Labs, USA \\
}

\begin{document}
\maketitle


\begin{abstract}
End-to-end approaches to autonomous driving have high sample complexity and are difficult to scale to realistic urban driving. Simulation can help end-to-end driving systems by providing a cheap, safe, and diverse training environment. Yet training driving policies in simulation brings up the problem of transferring such policies to the real world. We present an approach to transferring driving policies from simulation to reality via modularity and abstraction. Our approach is inspired by classic driving systems and aims to combine the benefits of modular architectures and end-to-end deep learning approaches. The key idea is to encapsulate the driving policy such that it is not directly exposed to raw perceptual input or low-level vehicle dynamics. We evaluate the presented approach in simulated urban environments and in the real world. In particular, we transfer a driving policy trained in simulation to a 1/5-scale robotic truck that is deployed in a variety of conditions, with no finetuning, on two continents.
\end{abstract}



\section{Introduction}

Autonomous navigation in complex environments remains a major challenge in robotics.
One important instantiation of this problem is autonomous driving.
Autonomous driving is typically addressed by highly engineered systems that comprise up to a dozen or more subsystems~\cite{Franke2017,Levinson2011,Paden2016,Urmson2008}.
Thousands of person-years are being invested in tuning these systems and subsystems. Yet the problem is far from solved, and the best existing solutions rely on HD maps and extensive sensor suites, while being relatively limited in the range of environmental and traffic conditions that can be handled.

End-to-end deep learning methods aim to substitute laborious hand-engineering by training driving policies directly on data~\cite{Pomerleau1988,LeCun2005driving,Bojarski2016nvidiadriving}.
However, these solutions are difficult to scale to realistic urban driving.
A huge amount of data is required to cover the full diversity of driving scenarios.
Moreover, deployment and testing are challenging due to safety concerns: the blackbox nature of end-to-end models makes it difficult to understand and evaluate the risks.

Simulation can help address these drawbacks of learning-based approaches.
In simulation, training data is abundant and driving policies can be tested safely.
Yet the use of simulation brings up a new challenge: transferring the learned policy from simulation to the real world.
This transfer is difficult due to the reality gap: the discrepancy in sensor readings, dynamics, and environmental context between simulation and the physical world.
Transfer of control policies for autonomous vehicles in complex urban environments is an open problem.

In this paper, we present an approach to bridging the reality gap by means of modularity and abstraction.
The key idea is to encapsulate the driving policy such that it is not directly exposed to raw perceptual input or low-level vehicle dynamics.
The architecture is organized into three major stages: perception, driving policy, and low-level control. First, a perception system maps raw sensor readings to a per-pixel semantic segmentation of the scene. Second, the driving policy maps from the semantic segmentation to a local trajectory plan, specified by waypoints that the car should drive through. Third, a low-level motion controller actuates the vehicle towards the waypoints. This is a more traditional architecture than end-to-end systems that map directly from image pixels to low-level control. We argue that this traditional architecture has significant benefits for transferring learned driving policies from simulation to reality. In particular, the policy operates on a semantic map (rather than image pixels) and outputs waypoints (rather than steering and throttle). We show that a driving policy encapsulated in this fashion can be transferred from simulation to reality directly, with no retraining or finetuning. This allows us to train the driving policy extensively in simulation and then apply it directly on a physical vehicle.

Both the perception system and the driving policy are learned, while the low-level controller can be either learned or hand-designed.
We train the perception system using publicly available segmentation datasets~\cite{Cordts2016}. The driving policy is trained purely in simulation.
Crucially, the driving policy is trained on the output of the actual perception system, as opposed to perfect ground-truth segmentation.
This allows the driving policy to adapt to the perception system's imperfections.

The combination of learning and modularization brings several benefits.
First and foremost, it enables direct transfer of the driving policy from simulation to reality. This is made possible by abstracting both the appearance of the environment (handled by the perception system) and the vehicle dynamics (handled by the low-level controller).
Second, the driving policy is still learned from data, and can therefore adapt to the complex noise characteristics of the perception system, which are not captured well by analytical uncertainty models. Lastly, the interfaces between the modules (semantic map, waypoints) are easy to analyze and interpret, which can help training and maintenance.

We evaluate the approach extensively on monocular-camera-based driving.
We experiment in simulated urban environments and in the real world, demonstrating both simulation-to-simulation and simulation-to-reality transfer.
In simulation, we transfer policies across environments and weather conditions and demonstrate that the presented modular approach outperforms its monolithic end-to-end counterparts by a large margin.
In the physical world, we transfer a policy from simulation to a 1/5-scale robotic truck, which is then deployed on a variety of roads (clear, snowy, wet) and in diverse environmental conditions (sunshine, overcast, dusk) on two continents.
The supplemental video demonstrates the learned policies:~\url{https://youtu.be/BrMDJqI6H5U}.

\section{Related Work}

\mypara{Transfer from simulation to the real world.}
Transfer from simulation to the real world has been studied extensively in computer vision and robotics.
Synthetic data has been used for training and evaluation of perception systems in indoor environments~\cite{Zhang2016physically,Handa2016,McCormac2017} and driving scenarios~\cite{Ros:2016,Gaidon:2016,Mayer:2016,Richter:2016,Skinner2016, JohnsonRoberson2017matrix,Tsirikoglou2017,Alhaija2017}.
Direct transfer from simulation to reality remains difficult even given high-fidelity simulation~\cite{McCormac2017,Richter:2016,Tsirikoglou2017,Zhang2016physically}, although successful examples exist for tasks such as optical flow estimation~\cite{Dosovitskiy2015} and object detection~\cite{JohnsonRoberson2017matrix,Hinterstoisser2017}.

Work on transfer of sensorimotor control policies has mainly dealt with manual grasping and manipulation.
A number of works employ specialized learning techniques and network architectures to facilitate transfer: for example, variants of domain adaptation~\cite{Tzeng2016,Gupta2017,Wulfmeier2017,Bousmalis2017,Pinto2017} and specialized architectures~\cite{Rusu2017,Zhang2017reaching}.
Working with depth maps instead of color images is known to simplify transfer~\cite{Viereck2017,Mahler2017dexnet, Mahler2017binpicking}.
When dealing with color images, domain randomization~\cite{Tobin2017,James2017,Sadeghi2017,Sadeghi2017:arxiv} enables direct sim-to-real generalization by maximizing the diversity of textures and occluders in simulation.
Other approaches tackle transfer via modularization in the context of manual grasping and pushing~\cite{Devin2017,Clavera2017}.
\citet{Clavera2017} propose an approach that is conceptually similar to ours in that a system is organized into a perception module, a high-level policy, and a low-level motion controller. However, their approach requires special instrumentation (AR tags) and was developed in the relatively constrained context of detecting and pushing an object with a robot arm against a uniform green-screen backdrop. \citet{Devin2017} also investigate the benefits of modularity in neural networks for control, but only evaluate in simulation, on tasks such as reaching towards and pushing colored blocks with a simulated arm against a uniform backdrop. In contrast, we develop policies that drive mobile robots outdoors, in dramatically more complex perceptual conditions. This requires different intermediate representations and modules.

Research on transfer of driving policies can be traced back to the seminal work of~\citet{Pomerleau1988}, who trained a neural network for lane following using synthetic road images and then deployed it on a physical vehicle. This early work is inspiring, but is restricted to rudimentary lane following.
More recently, \citet{Michels2005drivingrl} trained an obstacle avoidance policy in simulation and transferred it onto a small robotic vehicle via an intermediate depth-based representation. While their approach supports obstacle avoidance, it is not sufficient for urban driving, which requires more sophisticated perception and planning.

\citet{Sadeghi2017} perform transfer for UAV collision avoidance in hallways, using a high-quality 3D simulation with extensive domain randomization. This work is inspiring, but may be challenging to apply to outdoor urban driving, due to the high complexity of perception, planning, and control in this setting. Our work investigates a complementary approach that uses modularity to encapsulate the policy and abstract some of the nuisance factors that were tackled explicitly by Sadeghi and Levine via domain randomization.
\citet{Pan2017} use generative adversarial networks to adapt driving models from simulation to real images, and demonstrate improved performance on predicting human control commands, but do not validate their ideas with actual driving.

\mypara{Driving policies.}
Autonomous driving systems are commonly implemented via modular pipelines that comprise a multitude of carefully engineered components~\cite{Franke2017,Levinson2011,Paden2016,Urmson2008}.
The advantage of this modular design is that each component can be developed in isolation, and the system is relatively easy to analyze and interpret.
One downside is that such architectures can lead to accumulation of error.
Thus each component requires careful and time-consuming engineering.

\begin{figure*}
	\centering
       \includegraphics[width=\linewidth]{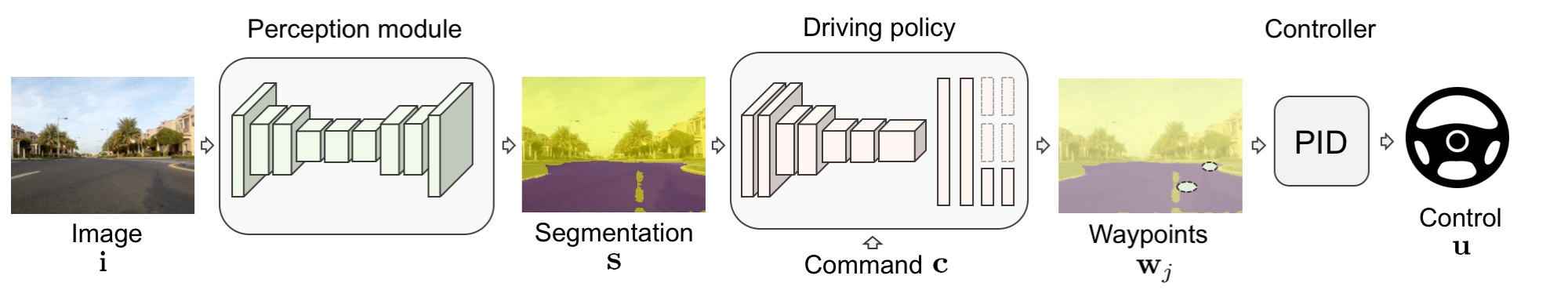}
    \vspace{-3mm}
    \caption{System architecture. The autonomous driving system comprises three modules: a perception module implemented by an encoder-decoder network, a command-conditional driving policy implemented by a branched convolutional network, and a low-level PID controller.}
	\label{fig:architechture}
\end{figure*}

Deep learning provides an appealing alternative to hand-designed modular pipelines.
Robotic vehicles equipped with neural network policies trained via imitation learning have been demonstrated to perform lane following~\cite{Pomerleau1988,Bojarski2016nvidiadriving}, off-road driving~\cite{LeCun2005driving,Silver2010navigation}, and navigation in simple urban environments~\cite{Codevilla2018}.
However, training these methods in the physical world requires expensive and time-consuming data collection.
Reinforcement learning in particular is known for its high sample complexity and is conducted mainly in simulation~\cite{Mnih2016a3c,Ebrahimi2017,Dosovitskiy2017}.
Application of deep reinforcement learning to real vehicles has only been demonstrated in restricted environments~\cite{Kahn2018}.

Finally, some approaches have explored the space between traditional modular pipelines and end-to-end learning.
\citet{Hadsell2009} train a perception system via self-supervised learning and use it in a standard navigation pipeline.
\citet{Chen:2015} decompose the driving problem into predicting affordances and then executing a policy based on these.
Concurrent work by \citet{Hong2018VirtualtoRealLT} explores a direction similar to our approach, but focuses on reinforcement learning in indoor environments.
Our work also aims to combine the best of traditional and end-to-end approaches.
In particular, we demonstrate that modularity allows transferring learned driving policies from simulation to reality.

\section{Method}

We address the problem of autonomous urban driving based on a monocular camera feed.
The overall architecture of the proposed driving system is illustrated in \figLabel \ref{fig:architechture}.
The system consists of three components: a perception module, a driving policy, and a low-level controller.
The perception module takes as input a raw RGB image and outputs a segmentation map.
The driving policy then takes this segmentation as input and produces waypoints indicating the desired local trajectory of the vehicle.
The low-level controller, given the waypoints, generates the controls: steering angle and throttle.
We now describe each of the three modules in detail.

\mypara{Perception.}
An image recorded by a color camera is affected by scene structure (e.g., layout of roads, buildings, cars, and pedestrians), surface appearance (materials, lighting), and properties of the camera.
The role of the perception system is to filter out nuisance factors and preserve the information needed for planning and control.
As the output representation for the perception system, we use a per-pixel binary segmentation of the image into ``road'' and ``not road'' regions.
It abstracts away texture, lighting, shading, and weather, leaving only a few factors of variation: the geometry of the road, the camera pose, and the shape of objects occluding the road.
Such segmentation contains sufficient information for following the road and taking turns, but it is abstract enough to support transfer.

We implement the perception system with an encoder-decoder convolutional network.
We train the network in supervised fashion on the binary road segmentation problem, with a cross-entropy loss.
The perception system is trained on the standard real-world Cityscapes segmentation dataset~\cite{Cordts2016}.
We find that networks trained on Cityscapes generalize sufficiently well both to the simulation and to the real environments we have experimented with.
Therefore, there is no additional effort needed on our side to generate training data for the segmentation network.
An analysis of the effect of the training dataset on the perception module is provided in the supplement.

We base the perception module on the ERFNet architecture~\cite{Romera2017}, which provides a favorable trade-off between accuracy and speed.
We further optimize the architecture for the task of binary segmentation to increase performance on an embedded platform. 
The exact architecture and training details are provided in the supplement.

\begin{wrapfigure}{r}{0.55\textwidth}
\vspace{-16pt}
\centering
\includegraphics[width=0.9\linewidth]{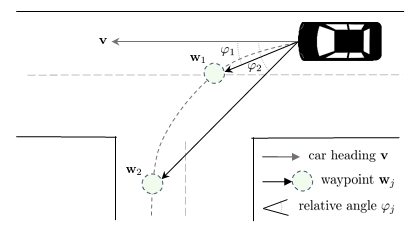}
    \caption{Waypoints are encoded by the distance to the vehicle and the relative angle to the vehicle's heading.}
	\label{fig:waypoint_encoding}
\end{wrapfigure}
\mypara{Driving policy.}
The driving policy takes as input the segmentation map produced by the perception system and outputs a local trajectory plan.
The plan is represented by waypoints that the vehicle has to drive through, illustrated in \figLabel \ref{fig:waypoint_encoding}.
At every frame, we predict two waypoints.
One would be sufficient to control steering, but the second can be useful for longer-term maneuvers, such as controlling the throttle ahead of a turn.
The waypoints $\mathbf{w}_j$ are encoded by the distance $r_j$ and the (oriented) angle $\varphi_j$ with respect to the heading direction $\mathbf{v}$ of the car:
\begin{equation}
\varphi_j = \angle(\mathbf{w}_j,\mathbf{v}), \qquad r_j = \norm{\mathbf{w}_j}.
\end{equation}
In our experiments we fix the distances to $r_1=5$ and $r_2=20$ meters for the two waypoints and only predict the angles $\varphi_1$ and $\varphi_2$.

We train the driving policy in simulation using conditional imitation learning (CIL)~\cite{Codevilla2018}~-- a variant of imitation learning that enables the driving policy to be conditioned on high-level commands, such as turning left or right at an upcoming intersection.
We now briefly describe CIL, and refer the reader to~\citet{Codevilla2018} for further details.

We start by collecting a dataset $\{\tuple{\obs_i,\cmd_i,\act_i}\}_{i=1}^N$ of observation-command-action tuples, from trajectories of an expert driving policy.
In our work, an observation can be an image or a segmentation map; the action can be either vehicle controls (steering, throttle) or waypoints; the command is a categorical variable indicating one of three high-level navigation instructions (\cmdleft, \cmdstraight, \cmdright) corresponding to driving left, straight, or right at the next intersection.
Given the training dataset, a function approximator $f$ with learnable parameters $\params$ is trained to predict actions from observations and commands:
\begin{equation}
\params^{*} = \argmin_{\params} \sum_i \ell(f(\obs_i,\cmd_i,\params),\, \act_i),
\end{equation}
where $\ell$ is a per-sample loss function, in our case mean squared error (MSE).
At test time the network can be guided by commands provided by a user or an automated high-level controller.

We use a deep network as the function approximator and adopt the branched architecture of~\citet{Codevilla2018}, with a shared convolutional encoder and a small fully-connected specialist network for each of the commands.
Compared to~\citet{Codevilla2018}, we change the inputs and the outputs of the network.
As input we provide the network with a binary road segmentation, encoded as a 2-channel image.
Instead of low-level controls, we train the network to output waypoints encoded by $\varphi_j$.
Further training details are provided in the supplement.

When training in simulation, a natural choice would be to use ground-truth segmentation as the input to the driving policy.
This may lead to good results in simulation, but does not prepare the network to deal with the noisy outputs of a real segmentation system in the physical world.
Therefore, in order to facilitate transfer, we train the network in simulation on noisy segmentation provided by a real perception system.
Interestingly, we have found that networks trained on the real-world Cityscapes dataset perform well both in the real world and in simulation.
Hence, instead of training a separate perception system in simulation, we use the same network as in the real world, trained on real data.
This introduces realistic noise and has the added benefit of direct transfer to the real world without even replacing the perception system.

In order to train the driving policy, we collect training data in simulation, using the CARLA platform~\cite{Dosovitskiy2017}.
Our dataset includes RGB images
from a front-facing camera and two additional cameras rotated by $30\deg$ to the left and to the right.
Making use of the capabilities provided by the simulator, we program an expert agent to drive autonomously based on privileged information: precise map and location of the ego-vehicle.
A global planner is used to randomly pick routes through a town and produce waypoints along the route.
A PID controller is used to follow these waypoints.
We collect training data in the absence of other agents~-- vehicles or pedestrians.
In order to increase the diversity of the dataset, the car is randomly initialized within the lane (not always in the center).
In total, we record $28$ hours of driving.
To improve the robustness of the learned policy, we follow~\citet{Codevilla2018} and introduce noise into the controls in approximately 20\% of the data.
We additionally randomize the camera parameters and perform data augmentation, as described in the supplement.

\mypara{Control.}
In order to convert the waypoints into control signals for the vehicle, we use a PID controller:
\begin{equation}
u(t) = K_p e(t) + K_i \int_{0}^{t} e(\tau) d\tau + K_d \frac{de(t)}{dt},
\end{equation}
where $u(t)$ is the control input, $e(t)$ is the error, and $K_p$, $K_i$, and $K_d$ are tuning parameters that correspond to proportional, integral, and derivate gains, respectively.

The low-level controls of our physical robotic truck are the throttle and the steering angle.
We use a PID controller for each of the controls: one for throttle (PID$_t$) and one for the steering angle (PID$_s$).
For PID$_t$, the error is the difference between the target speed and the current speed. For PID$_s$, the error is $\varphi_1$, the oriented angle between the viewing direction and the direction towards the first waypoint.

\begin{figure}[t]
	\centering
	\begin{tabular}{@{}c@{\hspace{1mm}}c@{\hspace{1mm}}c@{\hspace{1mm}}c@{}}
        & \small Maps & \small Weather 1  & \small Weather 2 \\
		\rotatebox{90}{\small ~~~~~~~~~Town 1} &
        \includegraphics[width=0.3\columnwidth]{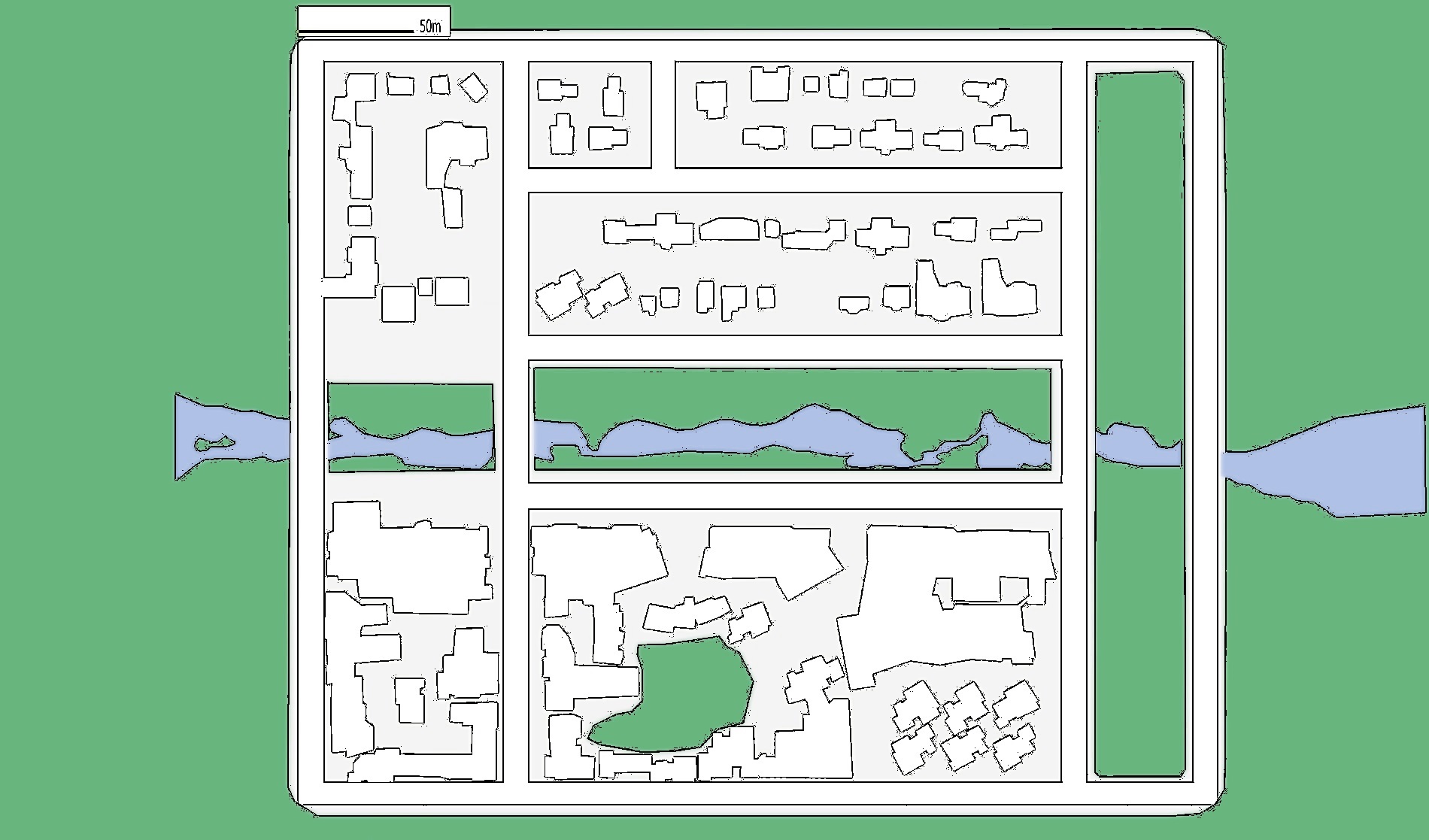} &
        \includegraphics[width=0.34\columnwidth]{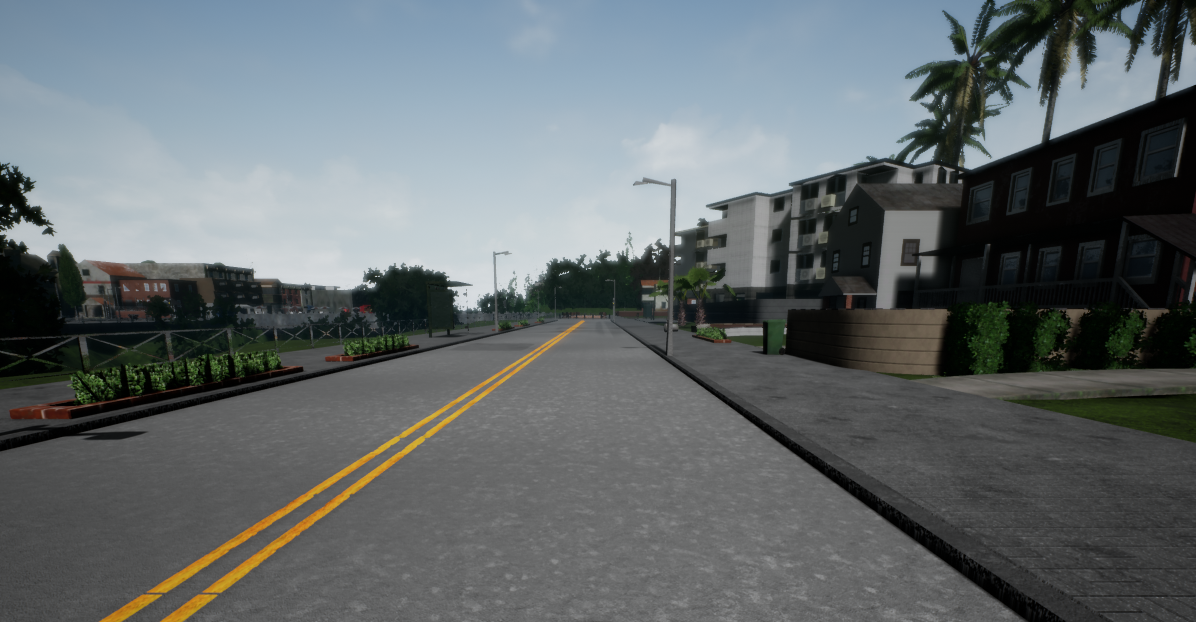} &
        \includegraphics[width=0.34\columnwidth]{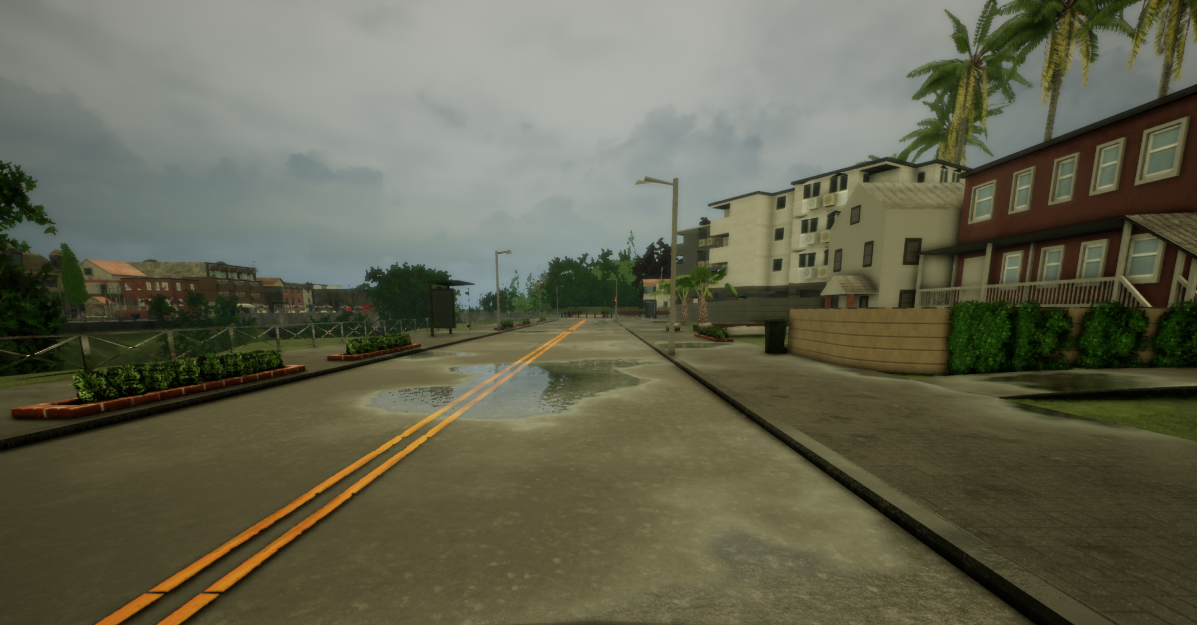} \\
		\rotatebox{90}{\small ~~~~~~~~~Town 2} &
        \includegraphics[width=0.3\columnwidth]{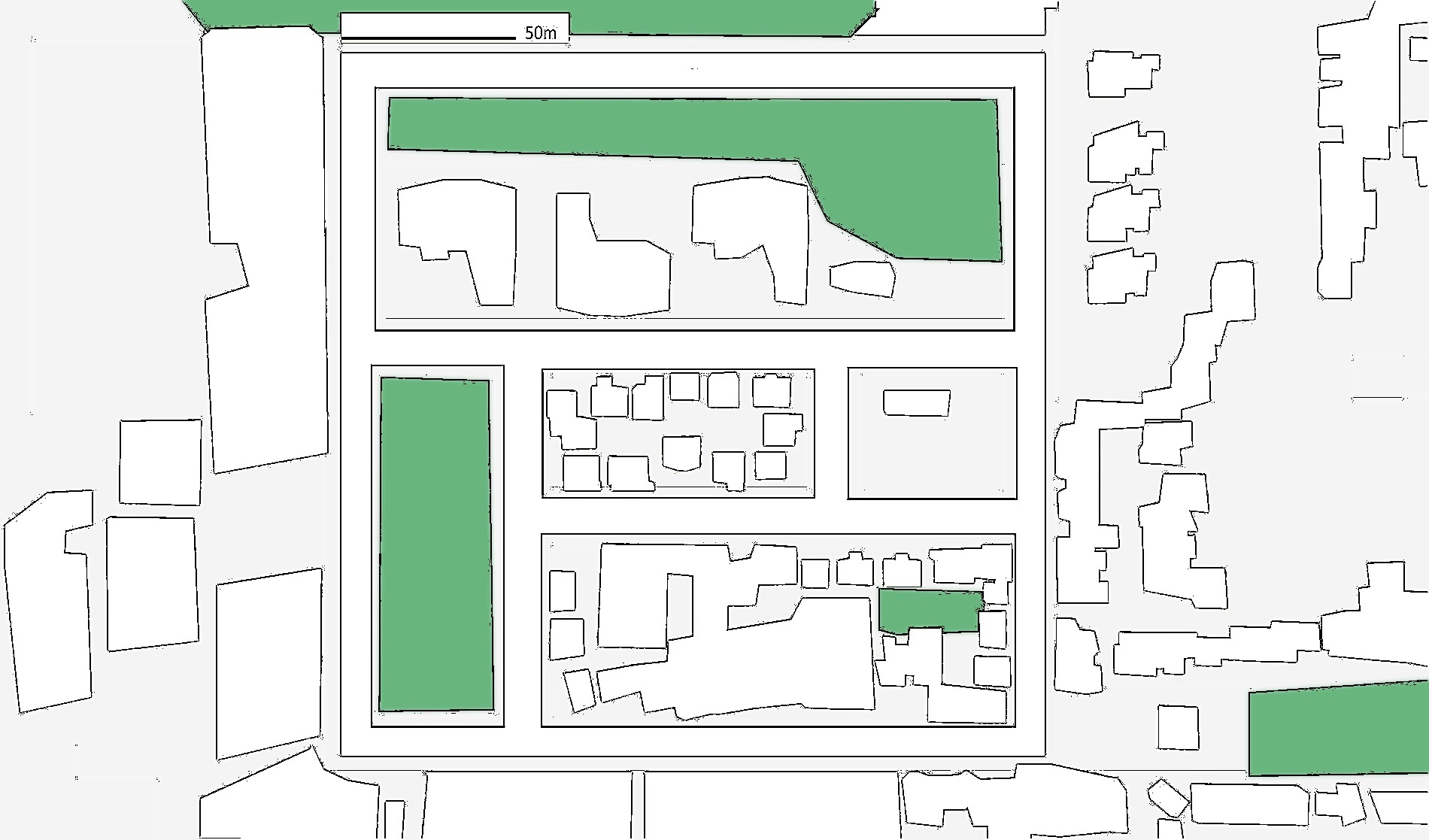} &
        \includegraphics[width=0.34\columnwidth]{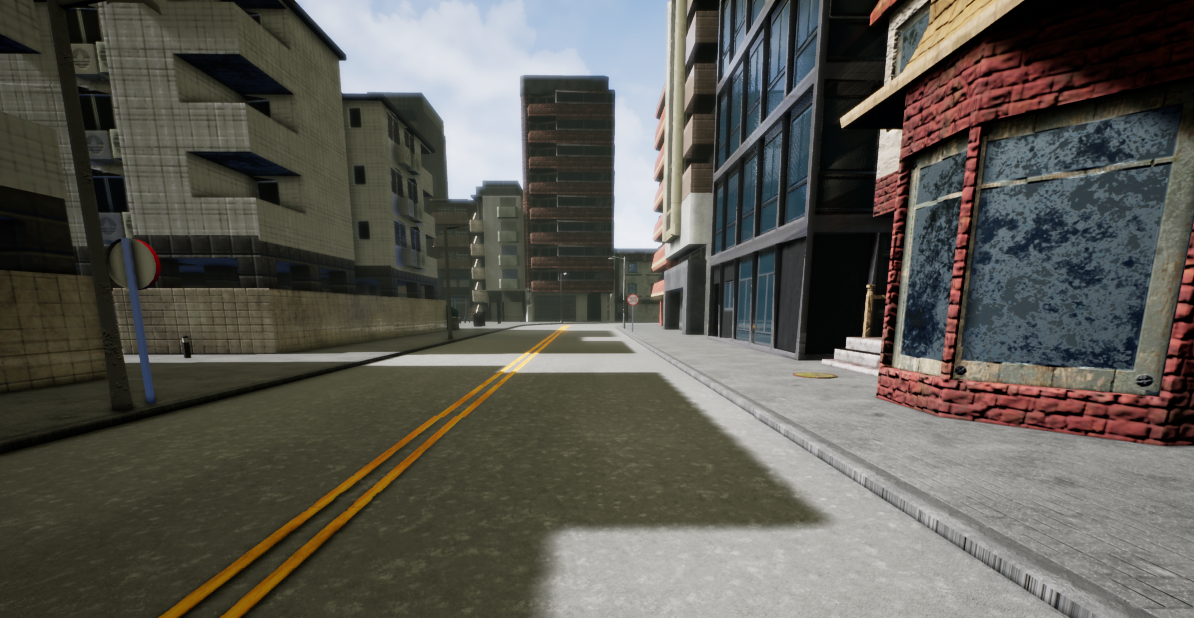} &
        \includegraphics[width=0.34\columnwidth]{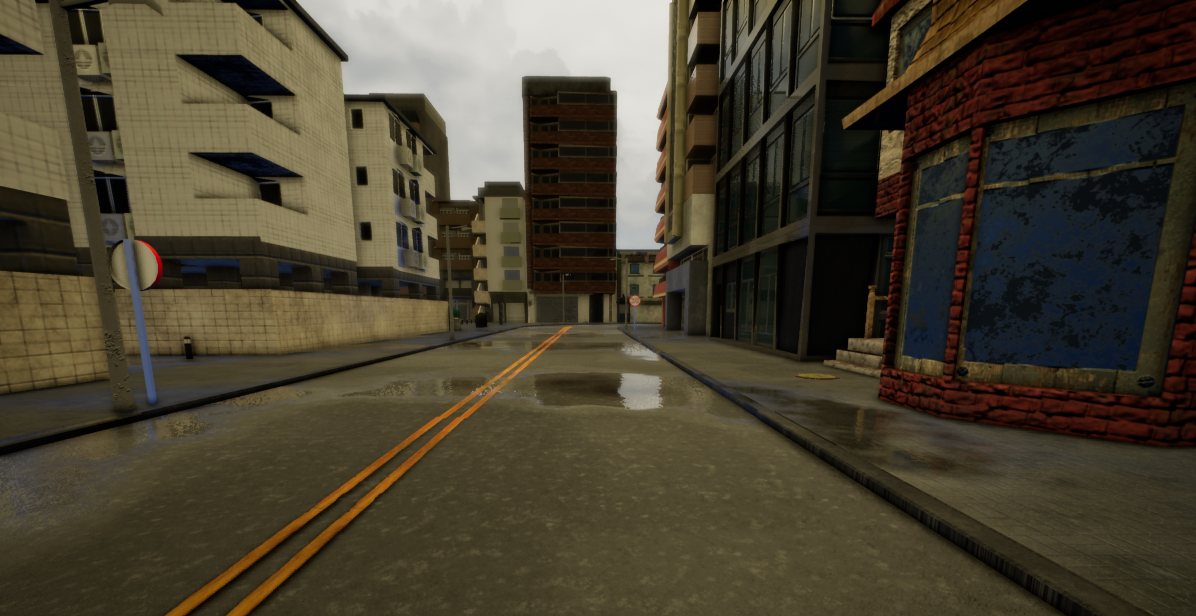}\\
	\end{tabular}
	\caption{Simulation environment. Maps of the two towns, along with example images that show the towns in two conditions: clear daytime (Weather 1) and cloudy daytime after rain (Weather 2). We use Town 1/Weather 1 during training. The other three combinations (Town 1/Weather 2, Town 2/Weather 1, and Town 2/Weather 2) are used to evaluate generalization in simulation. Note the significant visual differences between the towns and weather conditions.}
	\label{fig:carla}
\end{figure}

\section{System Setup}

We evaluate the presented approach in simulation and on a physical vehicle (a 1/5-scale truck).

\mypara{Simulation.}
We use CARLA, an open-source simulator for urban driving~\cite{Dosovitskiy2017}.
The simulator provides access to sensor data from the ego-vehicle, as well as detailed privileged information about the ego-vehicle and the environment.
The sensor suite of the vehicle can be easily specified by the user, and can include an arbitrary number of cameras returning color images, depth maps, or segmentation maps.
We make use of this flexibility when varying the camera FOV and position.
CARLA provides access to two towns: Town 1 and Town 2.
The towns differ in their layout, size, and visual style.
CARLA also provides multiple environmental conditions (combinations of weather and lighting). ~\figLabel \ref{fig:carla} illustrates our experimental setup.

\mypara{Physical system.}
We modified a 1/5 scale Traxxas Maxx truck to serve as an autonomous robotic vehicle.
The hardware setup is similar to~\citet{Codevilla2018}.
We equipped the truck with a flight controller (Pixhawk) running the APM Rover firmware which provides additional sensor measurements, allows for external control, and provides a failsafe system.
The bulk of computation, including the modular deep network, runs on an onboard computer (Nvidia TX2).
Attached to it is a USB camera that supplies the RGB images, and a FTDI adapter that enables communication with the Pixhawk. 
Given an image, the onboard computer predicts the waypoints and uses a PID controller to produce low-level control commands.
The steering angle and throttle are sent to the Pixhawk, which then converts them to PWM signals for the speed controller and steering servo.
While the car is driving, the driving policy can be guided by high-level command inputs (\cmdleft, \cmdstraight, \cmdright) through a switch on the remote control.
The complete system layout is provided in the supplement.

\section{Experiments}

\subsection{Driving in simulation}

We begin the driving experiments with a thorough quantitative evaluation on goal-directed navigation in simulation.
We use Town 1 in the clear daytime condition (Weather 1) for training.
To evaluate generalization, we benchmark the trained models in the same Town 1/Weather 1 condition and compare this to performance in three other conditions, which were not encountered during training: Town 1/Weather 2, Town 2/Weather 1, and Town 2/Weather 2 (Weather 2 is cloudy daytime after rain). See \figLabel \ref{fig:carla} for an illustration.

To evaluate driving performance we use a protocol similar to the navigation task of~\citet{Dosovitskiy2017}.
We select 25 start-goal pairs in each town, and perform a single goal-directed navigation trial for each pair.
In every trial, the vehicle is initialized at the start point and has to reach the goal point, given high-level commands from a topological planner.
We measure the percentage of successfully completed episodes.
This protocol is used to evaluate the performance of several driving policies.
\begin{itemize}
\setlength\itemsep{0.0mm}
\item
\textbf{Image to control (img2ctrl):}
Predicts low-level control directly from color images.
\item
\textbf{Image to waypoint (img2wp):}
Predicts waypoints directly from color images.
\item
\textbf{Segmentation to control (seg2ctrl):}
We pre-train the perception module on Cityscapes and fix it. We then train a driving policy to predict low-level control from segmentation maps produced by the perception module.
\item
\textbf{Segmentation to waypoint (ours):} Our full model predicts waypoints from segmentation maps produced by the perception module.
\end{itemize}

We additionally evaluate all these models trained with data augmentation.
We refer to these as \emph{img2ctrl+}, \emph{img2wp+}, \emph{seg2ctrl+}, and \emph{ours+}, respectively. We also compare to a variant of the domain randomization approach by \citet{Sadeghi2017}, which we refer to as \emph{img2wp+dr}.
Instead of randomizing the textures (which is not supported in CARLA), we uniformly sample from 12 different weather conditions when collecting the data, excluding Weather 2 used for testing.

\pgfplotstableread[row sep=\\,col sep=&]{
    interval & TrainT_TrainW & TrainT_TestW & TestT_TrainW & TestT_TestW \\
    img2ctrl    & 0.8  & 0  & 0.2 & 0  \\
    img2ctrl+    & 0.88  & 0  & 0.16 & 0  \\
    img2wp    & 0.72  & 0  & 0.36 & 0  \\
    img2wp+    & 0.92  & 0  & 0.44 & 0  \\
    img2wp+dr    & 0.64  & 0.76  & 0.64 & 0.16  \\ 
    seg2ctrl    & 0.64  & 0.32  & 0.24 & 0.08  \\
    seg2ctrl+ & 0.24 & 0.24 & 0.24 & 0.24 \\
    ours    & 0.96  & 0.92  & 0.68 & 0.28  \\
    ours+    & 0.84  & 0.84  & 0.8 & 0.56  \\
    }\wpdata

\begin{figure*}[t]
	\centering
    \vspace{-2mm}
\begin{tikzpicture}
    \fontsize{7}{8}
    \begin{axis}[
            ybar,
    		bar width=4pt,
    		enlarge x limits=0.1,
            width=0.9\textwidth,
            height=.28\textwidth,
            legend style={
            at={(1.2,1.0)},
            cells={align=left}},
            symbolic x coords={img2ctrl,img2ctrl+,img2wp,img2wp+,img2wp+dr,seg2ctrl,seg2ctrl+,ours,ours+},
            xtick=data,
            xtick style={draw=none},
            xticklabel style={text height=1ex},
            ymin=-0.01,ymax=1
        ]
        \addplot[blue!90!black,fill=blue!50!white!90!black] table[x=interval,y=TrainT_TrainW]{\wpdata};
        \addplot[red!80!black,fill=red!50!white!80!black] table[x=interval,y=TrainT_TestW]{\wpdata};
        \addplot[green!50!black,fill=green!50!white!50!black] table[x=interval,y=TestT_TrainW]{\wpdata};
        \addplot[yellow!50!black,fill=yellow!50!white!80!black] table[x=interval,y=TestT_TestW]{\wpdata};
        \addlegendentry{Town 1 \\ Weather 1}
        \addlegendentry{Town 1 \\ Weather 2}
        \addlegendentry{Town 2 \\ Weather 1}
        \addlegendentry{Town 2 \\ Weather 2}
    \end{axis}
\end{tikzpicture}
\vspace{-2mm}
    \caption{Quantitative evaluation of goal-directed navigation in simulation. We report the success rate over $25$ navigation trials in four town-weather combinations. The models have been trained in Town 1 and Weather 1. The evaluated models are: \emph{img2ctrl}~-- predicting low-level control from color images; \emph{img2wp}~-- predicting waypoints from color images; \emph{seg2ctrl}~-- predicting low-level control from the segmentation produced by the perception module; \emph{ours}~-- predicting waypoints from the segmentation produced by the perception module. Suffix `+' denotes models trained with data augmentation, and `+dr' denotes the model trained with domain randomization.}
    \label{fig:sim_results}
    \vspace{-4mm}
\end{figure*}
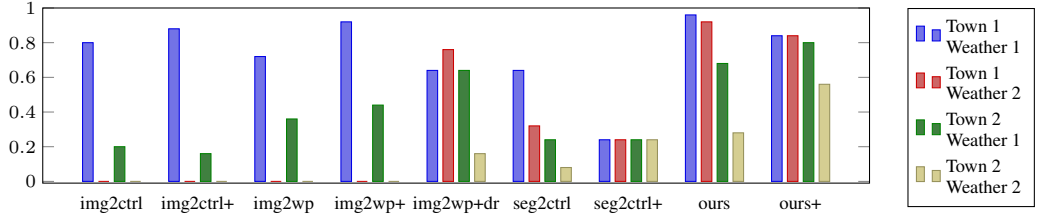

\figLabel \ref{fig:sim_results} presents the results of this comparison.
The most basic \emph{img2ctrl} control policy, trained end-to-end to predict low-level control from color images, drives fairly well under the training conditions.
It generalizes to Town 2 to some extent, but the success rate drops by a factor of $4$.
In the Weather 2 conditions, the model breaks down and does not complete a single episode.
Data augmentation slightly improves the performance in Town 1, but does not help generalization to Town 2.
Note that improved performance in the training environment (Town 1) when adding data augmentation is not unexpected: even when evaluated in the training environment, the network needs to generalize to previously unseen views.

The \emph{img2wp} model, trained to predict waypoints from color images, performs close to the \emph{img2ctrl} model under the training conditions.
However, it generalizes much better to Town 2, reaching roughly twice higher success rate than \emph{img2ctrl}.
This suggests that the waypoint representation is more robust to changes in the environment than the low-level control.
Data augmentation improves the performance in both towns.
Still, even with data augmentation, this model cannot generalize to the Weather 2 conditions and does not complete a single episode. Adding domain randomization (\emph{img2wp+dr}) enables the model to generalize to unseen weather to some extent.

The \emph{seg2ctrl} policy, predicting the low-level controls from segmentation produced by the perception module, is capable of non-trivial transfer to Weather 2. However, the overall performance is weak.

Finally, the presented approach (\emph{ours}), which predicts waypoints from segmentation, outperforms the baselines in the training environment and better generalizes to the test town.
Most importantly, it generalizes to the test weather condition in both towns better than the strongest baseline~-- domain randomization.
Data augmentation further boosts the generalization performance in the most challenging condition~-- Town 2/Weather 2~-- by a factor of 2.

\subsection{Driving in the physical world}
\label{sec:exp_real}
\pgfplotstableread[row sep=\\,col sep=&]{
    interval & success1 & success2 & SuccessRate \\
    img2wp    & 0 & 3 & 0.27 \\
    img2wp+    & 1 & 0 & 0.09 \\
    img2wp+dr    & -1 & 0 & 0.0 \\ 
    ours    & 2 & 7  & 0.82\\
    ours+    & 4 & 7 & 1.0   \\
    }\wpdata

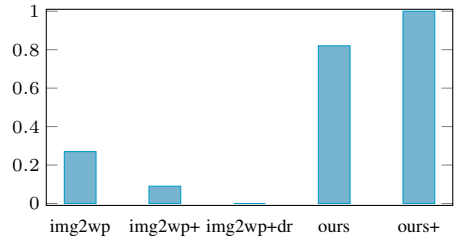
\begin{wrapfigure}{r}{0.45\textwidth}
\vspace{-12pt}
	\centering
\begin{tikzpicture}
    \fontsize{7}{8}
    \begin{axis}[
            ybar,
    		bar width=12pt,
    		enlarge x limits=0.1,
            width=0.5\textwidth,
            height=.3\textwidth,
			symbolic x coords={img2wp,img2wp+,img2wp+dr,ours,ours+},
            xtick=data,
            xtick style={draw=none},
            xticklabel style={text height=1ex},
            ymin=-0.01,ymax=1.01
        ]
       \addplot[cyan!80!black,fill=cyan!50!white!80!black] table[x=interval,y=SuccessRate]{\wpdata};

    \end{axis}
\end{tikzpicture}
\vspace{-2mm}
    \caption{Quantitative evaluation of road following in the real world. We report the average success rate over a total of $11$ navigation trials, with distance to be driven varying from $10$ to $50$ meters. Notation follows Figure~\ref{fig:sim_results}.}
    \label{fig:real_results}
\vspace{-6mm}
\end{wrapfigure}
\begin{figure*}
	\centering
    \setlength{\tabcolsep}{1mm}
    \begin{tabular}{cccc}
    	\includegraphics[width=0.235\linewidth]{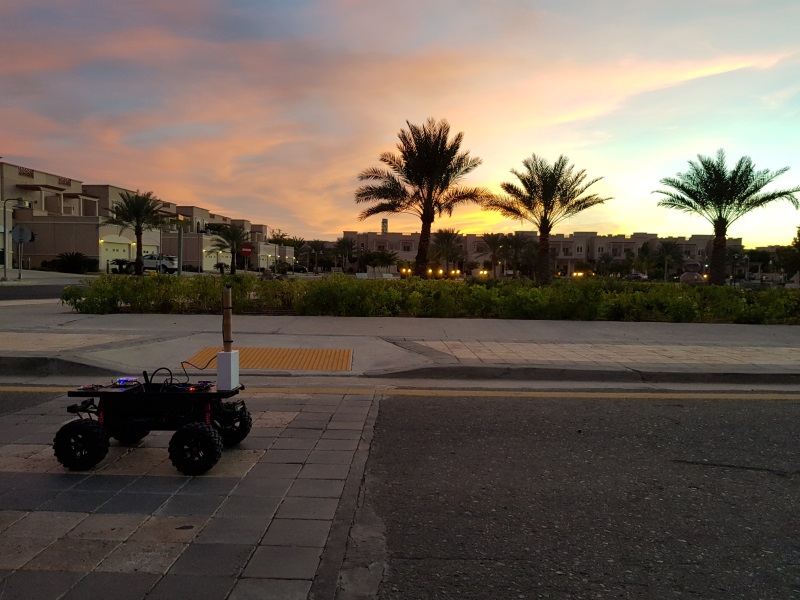} &
        \includegraphics[width=0.235\linewidth]{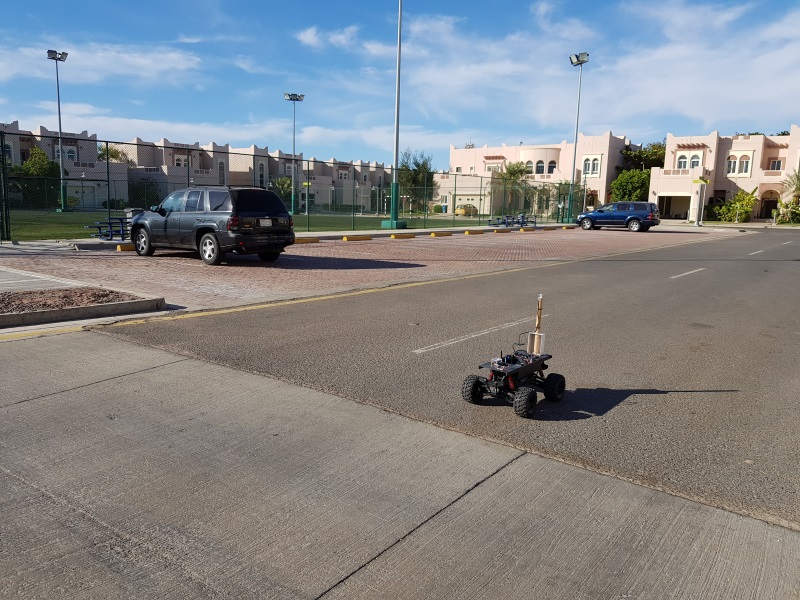} &
        \includegraphics[width=0.235\linewidth]{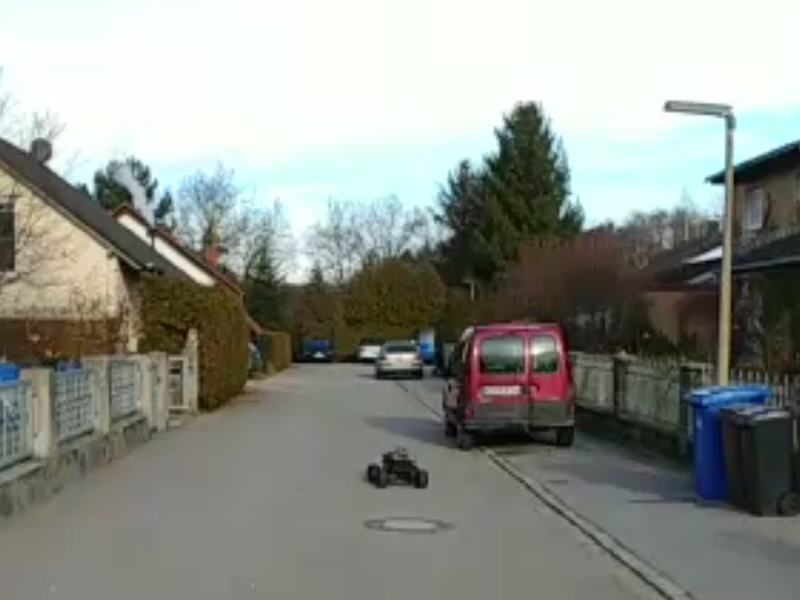} &
        \includegraphics[width=0.235\linewidth]{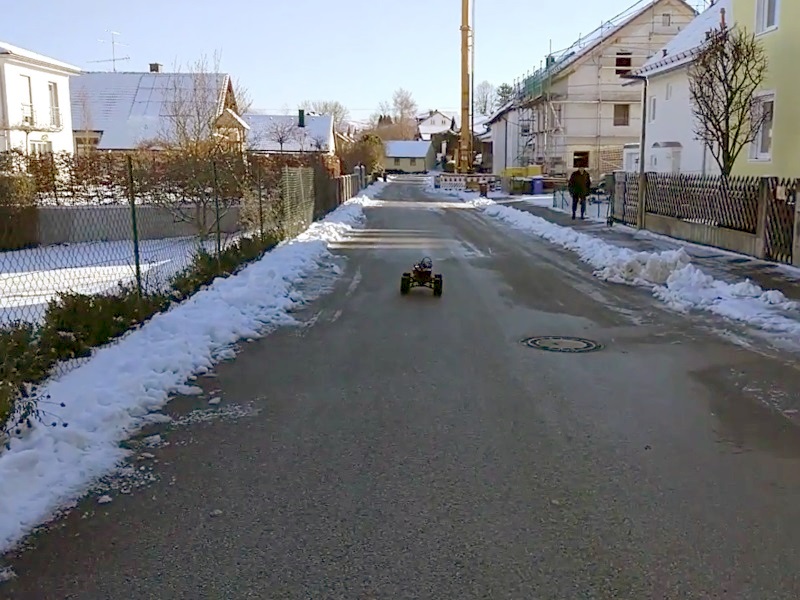}\\
	\end{tabular}
	\caption{Some of the environments the truck was tested in. Note the variation in scene structure, weather, and lighting. Qualitative results are shown in the supplementary video.}
	\label{fig:rctruck_environment}
    \vspace{-4mm}
\end{figure*}

We test the driving policy on the physical vehicle in multiple diverse and challenging environments.
Some of these are shown in \figLabel \ref{fig:rctruck_environment}.
Note the variation in the structure of the scene, the conditions of the road, and the lighting.
Qualitative driving results are shown in the supplementary video.

The road following capabilities of the learned policies are evaluated quantitatively.
We define several start locations on a road and measure the success rate of the vehicle reaching the end of the street from these locations.
We use $11$ locations spread over two geographic locations and different weather conditions, with distance to be driven varying from $10$ to $50$ meters.
The setup is illustrated in the supplement.

Figure \ref{fig:real_results} shows the performance of different models on this task.
In agreement with the sim-to-sim transfer results, the models trained directly on color images do not generalize well to the physical world, even with heavy augmentation or domain randomization.
Surprisingly, the basic \emph{img2wp} model is the most successful among these, reaching 27\% success rate, perhaps because of chance and the similarity between the clear sunny weather in CARLA used for training and the weather in some of the real world trials.
The proposed modular approach is able to complete 82\% of trials without data augmentation, and 100\% trials with data augmentation.
Based on these results, in what follows we only evaluate the best-performing model~-- our full system with data augmentation.

To measure navigation performance in the physical world, we have tested the vehicle on three routes. Detailed maps of these routes are provided in the supplement.
The routes include 7-8 turns each.
The vehicle is initialized in the beginning of a route and has to complete it, given navigational instructions sent by a human operator.
Commands are provided when a turn is roughly 5 meters ahead of the vehicle.
If a turn is missed, we record a missed turn and reset the vehicle to a position before the turn.
If the turn is missed again, we record it again and reset the vehicle after the turn. 

\begin{wraptable}{r}{0.6\textwidth}
\vspace{-2mm}
\centering
{\footnotesize
\vspace{-4mm}
\caption{Evaluation of our method on three long routes in an urban environment.}
\vspace{2mm}
\label{tbl:rctruck_results_kaust}
\begin{tabular}{ccccccc}
\toprule
       &  &          &        &                & \multicolumn{2}{c}{Infractions} \\ \cmidrule{6-7}
{Route\!\!\!\!\!\!} & & {Length}          & {Time}  & {Missed turns} & {Severe} & {Mild} \\ \midrule
1    &  & 1.0 km              & 4:12    & 1/7    & 0        & 2            \\
2    &  & 0.7 km              & 3:05    & 1/8    & 0        & 3            \\
3    &  & 1.1 km              & 5:08    & 2/8    & 1        & 5            \\ \bottomrule
\end{tabular}
}
\vspace{-2mm}
\end{wraptable}
The results are summarized in Table~\ref{tbl:rctruck_results_kaust}.
We report the number of missed turns, as well as the number of severe and mild infractions.
Severe infractions are those that require intervention, for instance a direct collision.
Mild infractions are those that the vehicle recovers from on its own~-- for instance, scraping the curb.
Table~\ref{tbl:rctruck_results_kaust} shows that the vehicle completes all three tracks while missing only a few turns (and completing all of them from the second try) and commits only one serious infraction requiring intervention (the vehicle drove onto the curb with two wheels and got stuck).
Although performance is not perfect, note that no real-world data was used to train the driving policy, other than the publicly available Cityscapes dataset that was used to train the perception system.
The driving policy itself was trained in simulation only.
To our knowledge, no previous methods demonstrated transfer to tasks and environments of this complexity.

\section{Conclusion}

We presented a modular deep architecture for autonomous driving, which integrates ideas from classic modular pipelines and end-to-end deep learning approaches.
Our model combines the benefits of both families of methods.
In comparison to a monolithic end-to-end network, the proposed architecture provides more flexibility.
Generalization to new environments (e.g., different weather, country, etc.) or transfer to new domains (e.g., simulation to physical world) can be achieved by appropriately tuning the perception module.
A human-interpretable interface between the modules simplifies analysis and debugging.
In comparison to classic driving pipelines, our driving policy is trained on the noisy output of a real perception module and can learn to be robust to complex error characteristics that are not captured by analytical uncertainty models.

While the results in our simplified scenario are promising, our approach needs to be further extended to make it useful for real autonomous vehicles.
First, although road segmentation is sufficient in some driving scenarios, it misses some important information such as lane markings, traffic signs, the state of traffic lights and dynamic obstacles (e.g., other vehicles and pedestrians). These can be added to the perception system or as separate inputs to the driving policy.
Second, the low-level controller we used is quite simple and future work can experiment with more complex controllers such as model predictive control or a learned control network.
Sophisticated controllers could be used to optimize the passengers' driving experience.
Third, training the driving policy in simulation enables the use of data-hungry learning algorithms such as reinforcement learning, which can be explored in this setting.
We see all of these as exciting directions for future research.



\mypara{Acknowledgments.~} This work was partially supported by the King Abdullah University of Science and Technology (KAUST) Office of Sponsored Research.

{\small
\setlength{\bibsep}{4pt plus 0.3ex}
\bibliographystyle{corlabbrvnat}
\bibliography{references}  
}

\clearpage
\appendix 
\section{Segmentation} \label{sec:exp_segm}

We evaluate the generalization of segmentation networks trained on different datasets.
To this end we collect and annotate a small real-world dataset in an urban area not overlapping with the areas used for driving experiments.
We split the dataset into a training set containing 70 images and a test set containing 36 images.
We evaluate the networks in simulation and on the test set of the small real-world dataset.
For training, we use standard public datasets~-- CamVid~\cite{Brostow2008}, Cityscapes~\cite{Cordts2016}, Berkeley Driving~\cite{Xu2017}, and a combination of all these~-- as well as our small dataset and data collected in simulation.

\begin{table}[!h]
\centering
{\small
\caption{Performance of the segmentation network in simulation and in the real world, when trained on different datasets. We report mean IoU (higher is better) and rank (lower is better) for each train-test combination, as well as the average rank across the two test datasets.}
\label{tbl::erfnet_miou}
\begin{tabular}{lcccc}
\toprule
                       & & \multicolumn{3}{c}{Testing}                 \\ \cmidrule{3-5}
 Training set          & & Simulated & Real & Avg. rank        \\ \midrule
 Simulation            & & 97.5 (1)    & 56.7 (6)       & 3.5          \\
 Physical World        & & 70.1 (6)    & 77.4 (3)       & 4.5          \\
 CamVid                & & 79.1 (5)    & 85.1 (2)       & 3.5          \\
 Cityscapes            & & 92.2 (2)    & 87.4 (1)       & \textbf{1.5} \\
 Berkeley              & & 87.3 (4)    & 75.6 (5)       & 4.5          \\
 All                   & & 91.2 (3)    & 76.6 (4)       & 3.5          \\ \bottomrule
\end{tabular}
}
\end{table}

The results are shown in Table \ref{tbl::erfnet_miou}.
As expected, a network trained in simulation works very well in simulation but does not generalize to the real world. 

\begin{figure*}[!h]
	\centering
    {
    \small
    \setlength{\tabcolsep}{0.5mm}
    \begin{tabular}{cccc}
        Simulation & Simulation & Real world & Real world \\
        \includegraphics[width=0.24\linewidth]{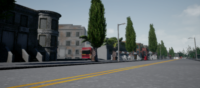} &
		\includegraphics[width=0.24\linewidth]{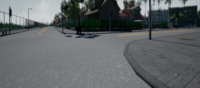} &
        \includegraphics[width=0.24\linewidth]{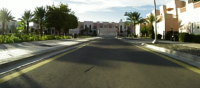} &
        \includegraphics[width=0.24\linewidth]{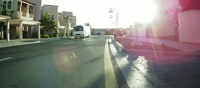} \\
        \includegraphics[width=0.24\linewidth]{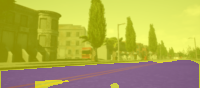} &
		\includegraphics[width=0.24\linewidth]{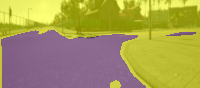} &
        \includegraphics[width=0.24\linewidth]{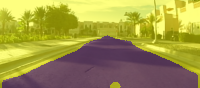} &
        \includegraphics[width=0.24\linewidth]{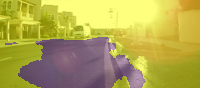} \\
	\end{tabular}
    }
	\caption{Sample outputs of the segmentation network trained on Cityscapes and tested in simulation and in the real world. The images are at the resolution used by the network~-- $200 \timess 88$ pixels. The network works well in typical scenes both in simulation and in the real world, but accuracy drops under complex lighting conditions and in unusual situations.}
	\label{fig:segm}
\end{figure*}

Interestingly, a network trained on Cityscapes generalizes to our validation data far better than other networks.
We attribute this primarily to the size and diversity of Cityscapes: more than 20K annotated images (including coarse annotations) from dozens of cities.
All following experiments use the segmentation network trained on Cityscapes for the perception system. Qualitative results are shown in Figure~\ref{fig:segm}. Note that while the results are quite good, they are far from perfect. Our learned driving policy is able to adapt to these imperfections, but it is likely that a better perception module could allow for even better driving performance.

We next compare the ERFNet-Fast architecture to the original ERFNet in terms of accuracy and runtime.
We train on Cityscapes and evaluate on our real-world test set.
Both networks are given $200 \timess 88$ pixel images, pre-loaded into RAM so as to make sure that only network execution time is benchmarked, without data input/output.
ERFNet-Fast achieves a mean IoU of $84.6\%$ while running at $25$ frames per second on the embedded platform.
The original ERFNet achieves a mean IoU of $85.8\%$ at $17$ frames per second.
This demonstrates that our architecture is well-suited to the task at hand, roughly matching ERFNet in accuracy while running at $40\%$ higher frame rate.
In addition, ERFNet-Fast has $9$ times fewer parameters than ERFNet.

\subsection{Network architecture}
\begin{figure}[h]
	\centering
	\begin{tabular}{@{}c@{\hspace{5mm}}c@{}}
        \includegraphics[width=0.35\columnwidth]{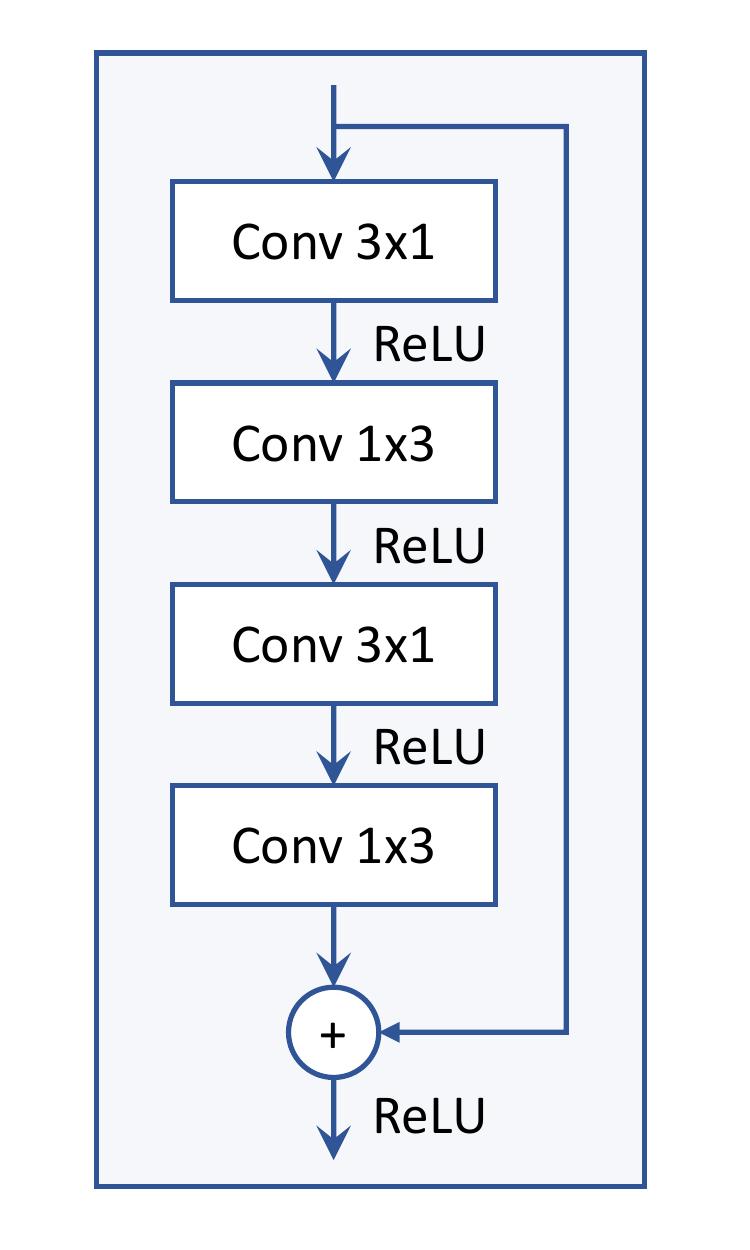} &
		\includegraphics[width=0.4\columnwidth]{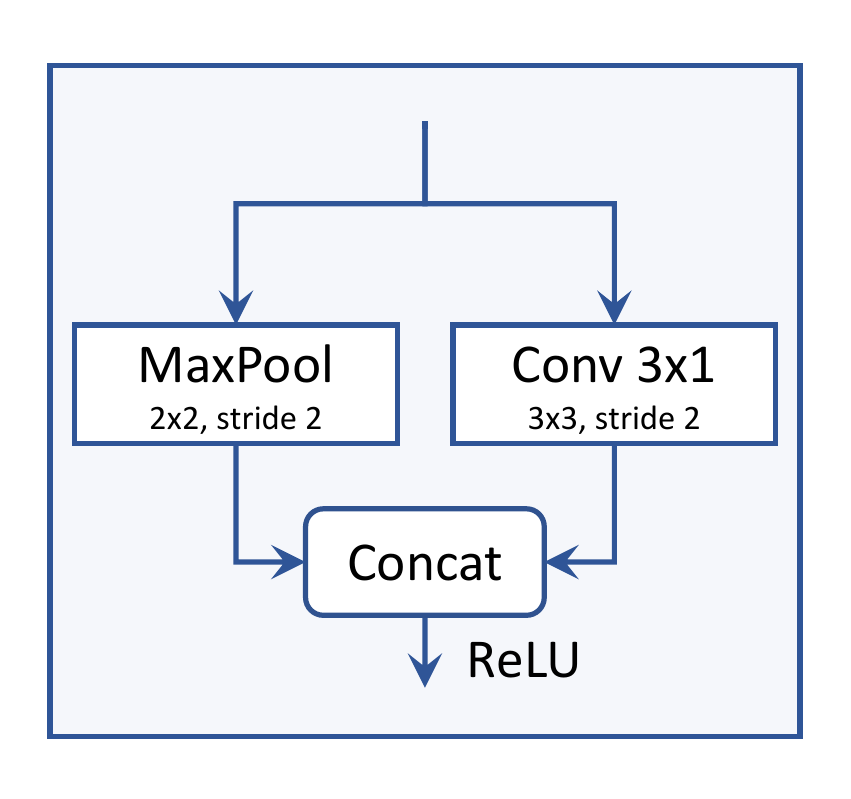} \\
		\small Non-bottleneck 1D & \small Downsampler
	\end{tabular}
	\caption{The ERFNet~\cite{Romera2017} building blocks, used in our architecture.}
	\label{fig:erfnet_blocks}
\end{figure}

The architecture of the segmentation network is shown in Table~\ref{tbl::erfnet}.
We use the modules from ERFNet~\cite{Romera2017} as building blocks.
These are shown in \figLabel~\ref{fig:erfnet_blocks}.
The architecture of the driving policy network is identical to~\citet{Codevilla2018}.

\begin{table}[h]
\centering
\caption{ErfNet-Fast architecture, used as perception module in our method.}
\label{tbl::erfnet}
\begin{tabular}{llcc}
\toprule
\textbf{Layer} & \textbf{Type}              & \textbf{out channels} & \textbf{out resolution} \\ \midrule
1              & Downsampler block          & 16             & $100 \timess 44$           \\
2-6            & 5 $\timess$ Non-bt-1D              & 16             & $100 \timess 44$           \\ \midrule
7              & Downsampler block          & 64             & $50 \timess 22$            \\
8              & Non-bt-1D (dilated 2)      & 64             & $50 \timess 22$            \\
9              & Non-bt-1D (dilated 4)      & 64             & $50 \timess 22$            \\
10             & Non-bt-1D (dilated 8)      & 64             & $50 \timess 22$            \\
11             & Non-bt-1D (dilated 16)     & 64             & $50 \timess 22$            \\ \midrule
12             & Deconvolution (upsampling) & 16             & $100 \timess 44$           \\
13-14          & 2 $\timess$ Non-bt-1D              & 16             & $100 \timess 44$           \\
15             & Deconvolution (upsampling) & 2              & $200 \timess 88$           \\ \bottomrule
\end{tabular}
\end{table}

\subsection{Training Details}
We implement all networks in TensorFlow~\cite{tensorflow} and train them using the Adam optimizer~\cite{Kingma2015adam}.
All networks operate on $200 \timess 88$ pixel images.

For the segmentation network, we set the initial learning rate to $0.001$ and reduce it to $0.0001$ after 100k iterations.
We use a batch size of $10$ and train until 200k iterations in total.
We use the method proposed by~\citet{Paszke2016} for class label balancing:
\begin{equation}
w_{c} = (\ln(p_{c}+ \gamma))^{-1}
\end{equation}
where $p_{c}$ is the average probability of class $c$ over the dataset and $\gamma=1.02$.

For the driving policy, we start with a learning rate of $0.0002$ and reduce it by a factor of 2 every 50k iterations until 250k. We train all models with a batch size of $120$ until 500k iterations. Both waypoints are weighted equally in the loss function.

\section{Data randomization}
\subsection{Camera parameters}
We vary the parameters of the camera when recording the dataset: the field of view (FOV) and the mounting position (height and orientation).
This is crucial for effective transfer, since otherwise the driving policy overfits to the specific camera being used during training.
In simulation it is easy to collect data using a variety of camera positions and views.
When recording the training data, we used cameras with 7 different FOVs, 3 different positions along the z-axis (50cm,100cm,150cm), and 3 different tilt angles ($-5\deg$, $0\deg$, $5\deg$).

\subsection{Data Augmentation}
To reach better generalization, in some of the training runs we regularize networks using data augmentation. We perform data augmentation at the image level: while training the driving policy, we randomly perturb the hue, saturation, and brightness of images that are fed to the perception module, and perform spatial dropout (that is, we set a random subset of input pixels to black).
Performing augmentation on the RGB images, not the segmentation maps themselves, produces more realistic variation in the segmentation maps. Further details on data augmentation are provided in the supplement.
We found that these additional regularization measures are crucial for achieving transfer to the real world.

When training segmentation networks we randomly perturb the input images as follows (assuming the pixel values are scaled between 0 and 1):
\begin{itemize}
\item Brightness: add a random number drawn uniformly from the interval $(-0.12,\, 0.12)$.
\item Saturation: multiply by a random factor drawn uniformly from the interval $(0.5,\, 1.5)$.
\item Hue: add a random number drawn uniformly from the interval $(-0.2,\, 0.2)$.
\item Contrast: multiply by a random factor drawn uniformly from the interval $(0.5,\, 1.5)$.
\end{itemize}

When training driving policies, we randomly perturb the input RGB images.
Depending on the architecture variant, these RGB images are being either fed directly to the driving policy, or to the perception module, which produces the segmentation map fed to the driving policy.
We use the following perturbations (assuming the pixel values are scaled between 0 and 1; for each perturbation, if it is applied, then it is applied with 50\% probability to each of the channels):
\begin{itemize}
\item Gaussian blur: with 5\% probability, apply Gaussian blur with standard deviation sampled uniformly from the interval $(0,\,1.3)$
\item Additive Gaussian noise: with 5\% probability, add Gaussian noise with standard deviation sampled uniformly from the interval $(0,\,0.05)$
\item Spatial dropout: with 5\% probability, set $d$ percent of RBG values to zero, with $d$ sampled uniformly from the interval $(0,\,0.1)$
\item Brightness additive: with 10\% probability, add to each of the channels a value sampled uniformly from the interval $(-0.08,\,0.08)$
\item Brightness multiplicative: with 20\% probability, multiply each of the channels by a value sampled uniformly from the interval $(0.25,\,2.5)$
\item Contrast multiplicative: with 5\% probability, multiply the contrast by a value sampled uniformly from the interval $(0.5,\,1.5)$
\item Saturation multiplicative: with 5\% probability, multiply the saturation by a value sampled uniformly from the interval $(0,\,1)$
\end{itemize}

\section{Physical System Setup}
\begin{figure}[!htb]
	\centering
    \includegraphics[width=0.8\columnwidth]{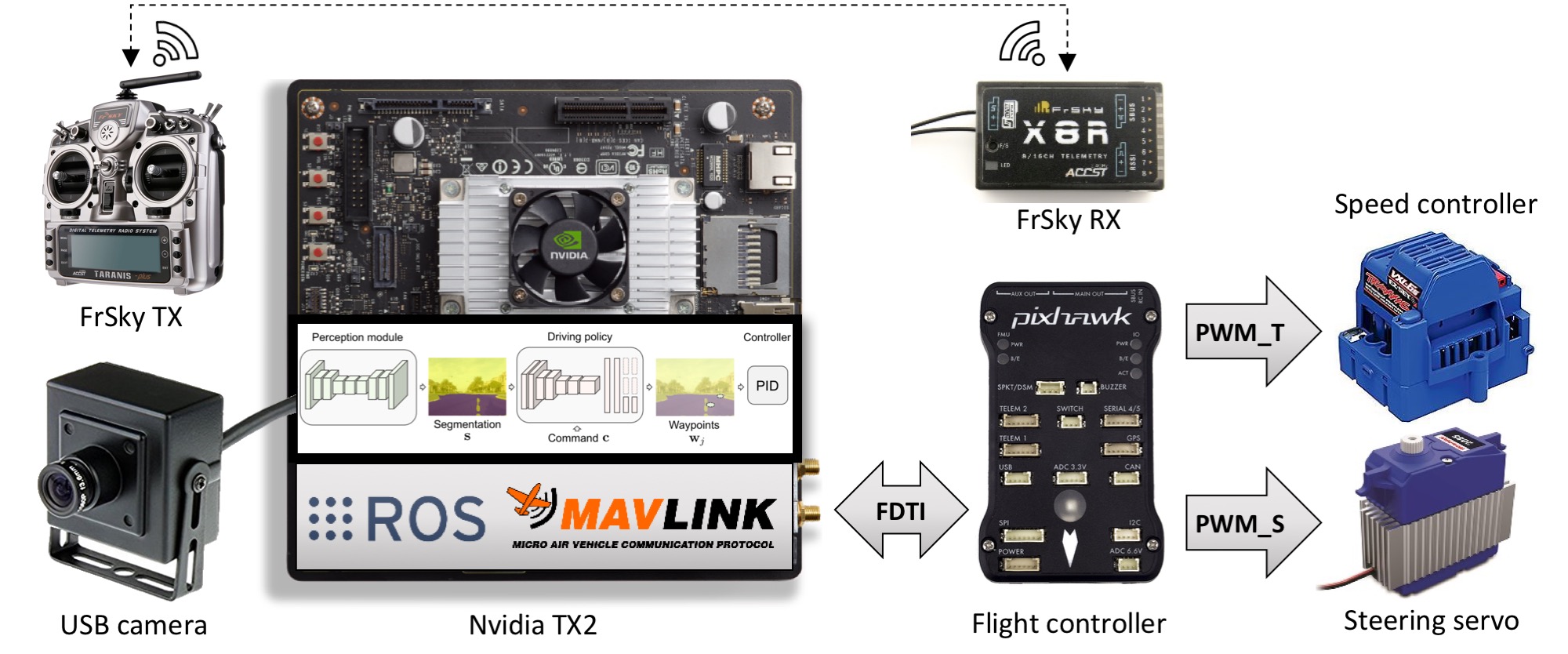}
	\caption{Hardware setup for the robotic vehicle.}
	\label{fig:rctruck_setup}
\end{figure}

Figure \ref{fig:rctruck_setup} shows the setup of our physical system. All of the components except for the remote control are mounted to the RC truck. The operator can toggle the autonomous driving mode from the remote control.

At runtime, given an image, the onboard computer predicts the waypoints and uses a PID controller to convert them to low-level control commands.
We found the speed estimate provided by the flight controller to be very unreliable, therefore we fix the throttle in our experiments so that the car drives at a constant speed of approximately 3 m/s.
In order to compute the steering angle, we use a PID controller with coefficients $K_p=0.8$,  $K_i=0$, and $K_d=0$.
The steering angle and throttle are sent to the Pixhawk using the MAVROS package, which converts them to the low-level PWM signals for the speed controller and steering servo.
While the car is driving, the driving policy can be guided by high-level command inputs (\cmdleft, \cmdstraight, \cmdright) through a switch on the remote control.

We have experimented with two ways of mounting the camera.
In one, the camera is mounted low under a protective shell, to protect the electronics from the changing weather conditions.
We use this setup for lane-following experiments.
In this setup, the field of view of the camera is very restricted, which can lead to missed turns.
We therefore experimented with another configuration~-- mounting the camera higher, so as to increase the field of view.
We use this setup for complex navigation involving turns, in clear weather.

\subsection{Experimental Setup}
\begin{figure}[!htb]
	\centering
	\begin{tabular}{@{\hspace{10mm}}c@{\hspace{15mm}}c@{\hspace{15mm}}c@{\hspace{10mm}}c}
        \multicolumn{4}{c}{\includegraphics[width=0.8\linewidth]{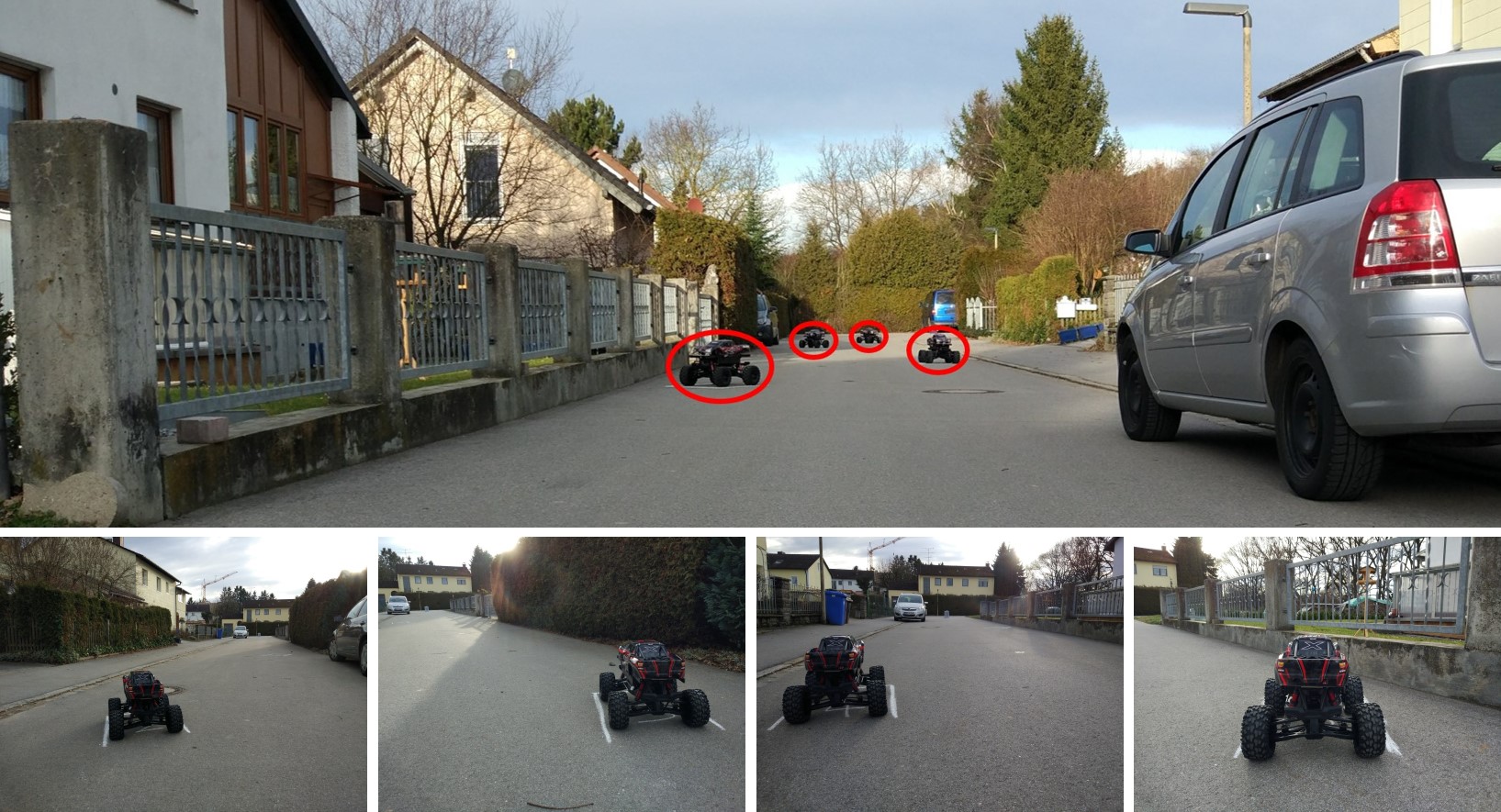}} \\
		\small Position 1 & \small Position 2 & \small Position 3 & \small Position 4
	\end{tabular}
	\caption{Controlled evaluation of road following performance in environment 1. Top: a composite image showing four starting positions as seen from the finish line. Bottom: close-ups of the four starting positions. The positions are clearly marked for consistency across episodes.}
	\label{fig:rctruck_test_setup}
\end{figure}

\begin{figure}[!htb]
	\centering
	\begin{tabular}{@{\hspace{8mm}}c@{\hspace{9mm}}c@{\hspace{10mm}}c@{\hspace{10mm}}c@{\hspace{7mm}}c}
        \multicolumn{5}{c}{\includegraphics[width=0.8\linewidth]{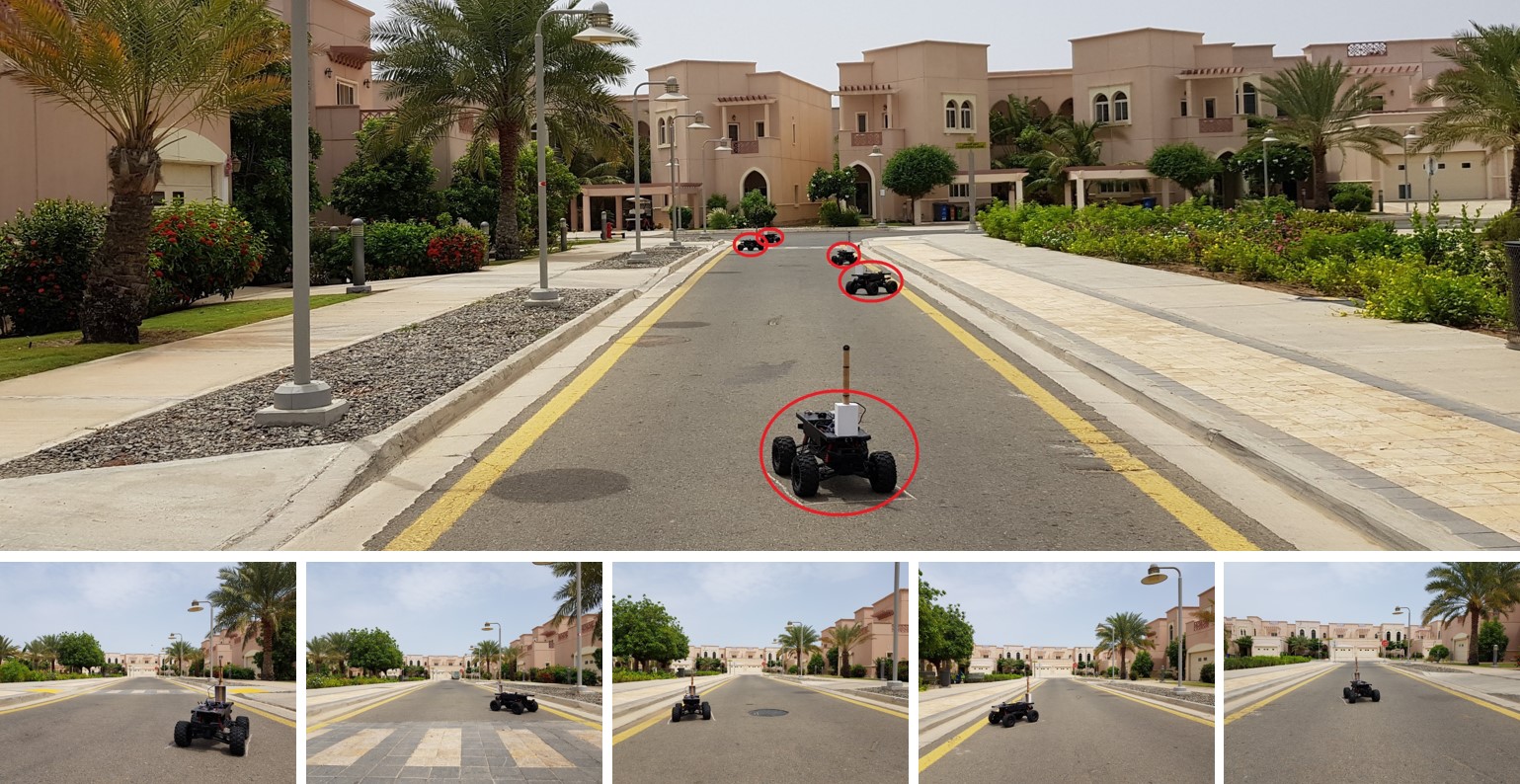}} \\
		\small Position 1 & \small Position 2 & \small Position 3 & \small Position 4 & \small Position 5
	\end{tabular}
	\caption{Controlled evaluation of road following performance  in environment 2. Top: a composite image showing five starting positions as seen from the finish line. Bottom: close-ups of the five starting positions. The positions are clearly marked for consistency across episodes.}
	\label{fig:rctruck_test_kaust}
\end{figure}

Figure \ref{fig:rctruck_test_setup} shows the $4$ starting positions in the first of the two environments for the lane-following experiment. Figure \ref{fig:rctruck_test_kaust} shows the $5$ starting positions in the second environment. In addition, in the second environments we executed two more runs starting about 10 meters before an intersection and the vehicle is commanded to turn left and right. 

\begin{figure}[!ht]
	\centering
	\begin{tabular}{c@{\hspace{1.5mm}}c@{\hspace{1.5mm}}c}
   		\includegraphics[height=4.5cm]{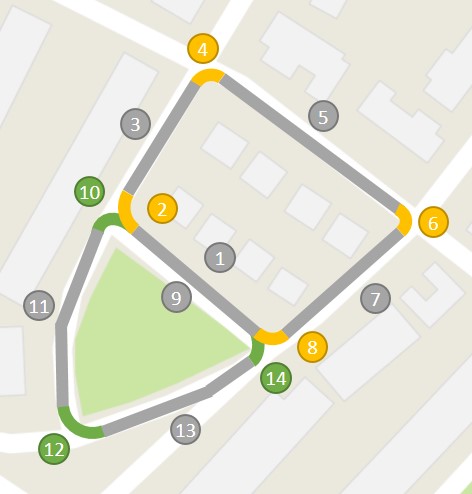} &
        \includegraphics[height=4.5cm]{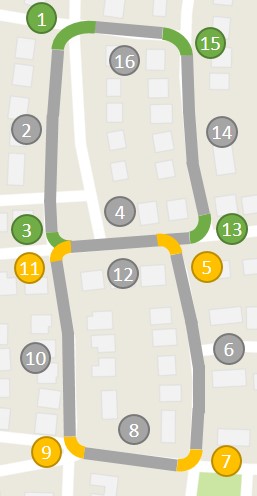} &
        \includegraphics[height=4.5cm]{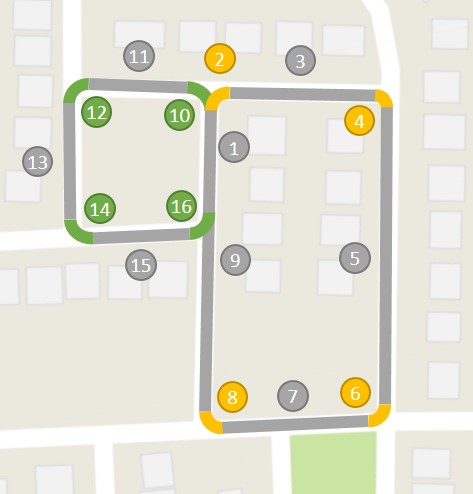} \\
		\small Route 1 & \small Route 2 & \small Route 3
	\end{tabular}
	\caption{Three routes used for evaluating the driving policy. Right turns marked in yellow, left turns in green.}
	\label{fig:rctruck_kaust}
\end{figure}

These basic experiments were used to determine the best model which should be able to follow the lane, recover from various position and able to do left and right turns. We then pick the best model and evaluate it on the more difficult navigation task where the vehicle has to complete several trajectories with various turns. Figure \ref{fig:rctruck_kaust} shows the maps for the more complex navigation experiments.

\end{document}